\documentclass[lettersize,journal]{IEEEtran}
\usepackage{amsmath,amsfonts}
\usepackage{amsthm}
\usepackage{algorithmic}
\usepackage{algorithm}
\usepackage{array}
\usepackage[caption=false,font=normalsize,labelfont=sf,textfont=sf]{subfig}
\usepackage{textcomp}
\usepackage{stfloats}
\usepackage{url}
\usepackage{verbatim}
\usepackage{natbib}
\usepackage{multirow}
\usepackage{graphicx}
\usepackage{tabularx}
\usepackage{caption}
\usepackage{subcaption}
\usepackage{booktabs}
\usepackage{xcolor}
\usepackage{xspace}
\usepackage{url}

\usepackage[edges]{forest}
\usepackage{multicol}
\usepackage{makecell}

\usepackage{verbatim}
\usepackage{tikz}
\usepackage{balance}

\newcommand{\hide}[1]{}

\definecolor{hidden-draw}{RGB}{20,68,106}
\definecolor{hidden-pink}{RGB}{255,245,247}
\definecolor{lightred}{RGB}{255, 204, 204}
\definecolor{lightgreen}{RGB}{224, 255, 225}
\definecolor{lightyellow}{RGB}{255, 241, 224}
\definecolor{lightpurple}{RGB}{225, 225, 255}
\definecolor{lightgray}{gray}{0.9}
\definecolor{text-red}{RGB}{255, 0, 0}
\definecolor{text-blue}{RGB}{0, 0, 255}
\definecolor{deep-purple}{RGB}{84, 74, 255}
\definecolor{deep-blue}{RGB}{0, 170, 238}
\definecolor{deep-green}{RGB}{63, 183, 4}

\newcommand{\txrd}[1]{\textcolor{text-red}{\textbf{#1}}}
\newcommand{\txbl}[1]{\textcolor{text-blue}{\textbf{#1}}}
\newcommand{\xs}[1]{\textcolor{text-blue}{#1}}

\renewcommand{\txrd}[1]{\textbf{#1}}
\renewcommand{\txbl}[1]{\underline{#1}}

\renewcommand{\xs}[1]{\textcolor{black}{#1}}

\usepackage{xspace}
\usepackage{fontawesome5}
\usepackage[colorlinks=true,linkcolor=blue,urlcolor=blue,citecolor=blue]{hyperref}  

\def\BibTeX{{\rm B\kern-.05em{\sc i\kern-.025em b}\kern-.08em
    T\kern-.1667em\lower.7ex\hbox{E}\kern-.125emX}}

\def\model{\textbf{MODE}}

\begin{document}

\title{MODE: Efficient Time Series Prediction with Mamba Enhanced by Low-Rank Neural ODEs}

\author{
    Xingsheng Chen,
    Regina Zhang,
    Bo Gao,
    Xingwei He,
    Xiaofeng Liu,\\
    Pietro Lio,
    Kwok-Yan Lam,
    Siu-Ming Yiu
    \thanks{This work has been submitted to the IEEE for possible publication. Copyright may be transferred without notice, after which this version may no longer be accessible. (Corresponding authors: Regina Zhang, Kwok-Yan Lam and Siu-Ming Yiu).}
    \IEEEcompsocitemizethanks{
        \IEEEcompsocthanksitem X. Chen is with the School of Computing and Data Science, The University of Hong Kong.
        \IEEEcompsocthanksitem R. Zhang is in the Department of Computing and Data Science, Nanyang Technological University.
        \IEEEcompsocthanksitem B. Gao is with the School of Information Engineering, Beijing Institute of Graphic Communication.
        \IEEEcompsocthanksitem X. He is in the Department of Computing and Data Science, The University of Hong Kong.
        \IEEEcompsocthanksitem X. Liu works at Yale university.
        \IEEEcompsocthanksitem P. Lio works at University of Cambridge.
        \IEEEcompsocthanksitem K. Lam works in Nanyang technological university.
        \IEEEcompsocthanksitem S. Yiu works at University of Hong Kong.
    }
}

\maketitle

\begin{abstract}
Time series prediction plays a pivotal role across diverse domains such as finance, healthcare, energy systems, and environmental modeling. However, existing approaches often struggle to balance efficiency, scalability, and accuracy, particularly when handling long-range dependencies and irregularly sampled data. To address these challenges, we propose \textbf{\model}, a unified framework that integrates Low-Rank Neural Ordinary Differential Equations (Neural ODEs) with an Enhanced Mamba architecture. As illustrated in our framework, the input sequence is first transformed by a Linear Tokenization Layer and then processed through multiple Mamba Encoder blocks, each equipped with an Enhanced Mamba Layer that employs Causal Convolution, SiLU activation, and a Low-Rank Neural ODE enhancement to efficiently capture temporal dynamics. This low-rank formulation reduces computational overhead while maintaining expressive power. Furthermore, a segmented selective scanning mechanism, inspired by pseudo-ODE dynamics, adaptively focuses on salient subsequences to improve scalability and long-range sequence modeling. Extensive experiments on benchmark datasets demonstrate that \textbf{\model} surpasses existing baselines in both predictive accuracy and computational efficiency. Overall, our contributions include: (1) a unified and efficient architecture for long-term time series modeling, (2) integration of Mamba's selective scanning with low-rank Neural ODEs for enhanced temporal representation, and (3) substantial improvements in efficiency and scalability enabled by low-rank approximation and dynamic selective scanning. The code and data used in this work are available at the following link:~\href{https://github.com/XsChen524/mode-mamba-ts}{https://github.com/XsChen524/mode-mamba-ts}.
\end{abstract}

\begin{IEEEkeywords}
Time series prediction, Neural ordinary differential equations (Neural ODEs), Low-rank approximation, Mamba architecture, Sequence modeling
\end{IEEEkeywords}

\section{Introduction}

\IEEEPARstart{T}{ime} series prediction~\cite{lim2021time,wen2022transformers,zhang2025autohformer,esling2012time,zhang2018time} is a fundamental task in machine learning and statistics, underpinning a wide range of real-world applications, including finance, healthcare, climate modeling, social networks~\cite{zheng2023cipl,zheng2024hycorec,zheng2024mitigating,zhang2025fldmamba,zheng2025cirec}, and urban forecasting~\cite{zhang2025efficient,zhang2024survey,zhang2025survey,zhangprompts,mahalakshmi2016survey}. The overarching objective is to learn temporal dependencies within sequential data in order to accurately forecast future observations. Despite significant progress, time series prediction remains challenging due to the complex characteristics of real-world data, which often exhibit nonlinear dynamics, long-range dependencies, and irregular sampling patterns. 

Traditional approaches, such as autoregressive models~\cite{ARIMA2} and recurrent neural networks (RNNs)~\cite{medsker2001recurrent}, typically struggle to capture long-term dependencies or complex temporal behaviors. Recent advances, particularly those based on Transformers~\cite{vaswani2017attention} and state-space models (SSMs)~\cite{auger2021guide}, have shown enhanced modeling capability for sequential data. However, these models are often limited by high computational costs, insufficient scalability, and difficulty in handling irregularly sampled time series effectively.

The fundamental challenges in time series forecasting can be summarized as follows. \textit{First}, long-range dependency modeling remains difficult, as the influence of distant observations on future outcomes spans extended time intervals. This is especially critical in domains like financial forecasting and climate modeling, where long-term trends can strongly influence predictive outcomes. \textit{Second}, irregularly sampled data, prevalent in practical applications such as healthcare, sensor networks, and event-driven systems, violates the uniform time-step assumption of conventional discrete-time models, often leading to degraded performance. \textit{Third}, computational efficiency becomes a critical bottleneck for long sequences and high-dimensional data, as the time and memory complexity of transformer-based or SSM-based models can scale quadratically with sequence length. Consequently, there is a pressing need for a predictive framework that effectively models temporal dependencies, naturally accommodates irregular sampling, and scales efficiently to large and complex datasets.

To address these challenges, we propose \textbf{\model}, a novel and unified framework that integrates Low-Rank Neural Ordinary Differential Equations (Neural ODEs) with the Mamba architecture. The design of \textbf{\model} directly targets the key limitations of existing methods. Specifically, the continuous-time property of Neural ODEs enables the model to process irregularly sampled data naturally without resorting to explicit imputation or interpolation, a crucial advantage for applications with sparse or uneven sampling such as healthcare and environmental monitoring. Furthermore, a low-rank approximation strategy is introduced to reduce the computational complexity of state transitions from $\mathcal{O}(d^2)$ to $\mathcal{O}(d \cdot r)$, where $r \ll d$, allowing for efficient modeling of high-dimensional sequences. In addition, the selective scanning mechanism guides the model’s attention toward the most informative segments of the sequence, further enhancing scalability and efficiency for long-term forecasting tasks. Together, these components make \textbf{\model} a robust, efficient, and scalable solution for modern time series prediction.

Our main contributions are summarized as follows:
\begin{itemize}
    \item \textbf{Unified and Efficient Framework for Long-Term Time Series Prediction:} We introduce a principled framework that seamlessly integrates Low-Rank Neural ODEs with the Mamba structure, enabling effective and adaptive modeling of complex temporal dynamics without manual feature engineering.
    
    \item \textbf{Selective Scanning for Scalable Temporal Modeling:} We design a segmented selective scanning mechanism that prioritizes salient temporal segments, substantially improving computational efficiency and scalability while preserving high predictive accuracy.
    
    \item \textbf{Comprehensive Empirical Evaluation:} Extensive experiments across multiple benchmark datasets demonstrate that \textbf{\model} achieves state-of-the-art performance compared to transformer-based and SSM-based baselines, confirming its robustness, scalability, and generalization capabilities across diverse time series prediction tasks.
\end{itemize}


\begin{figure*}[!t]
    \centering
\includegraphics[width=0.75\textwidth]{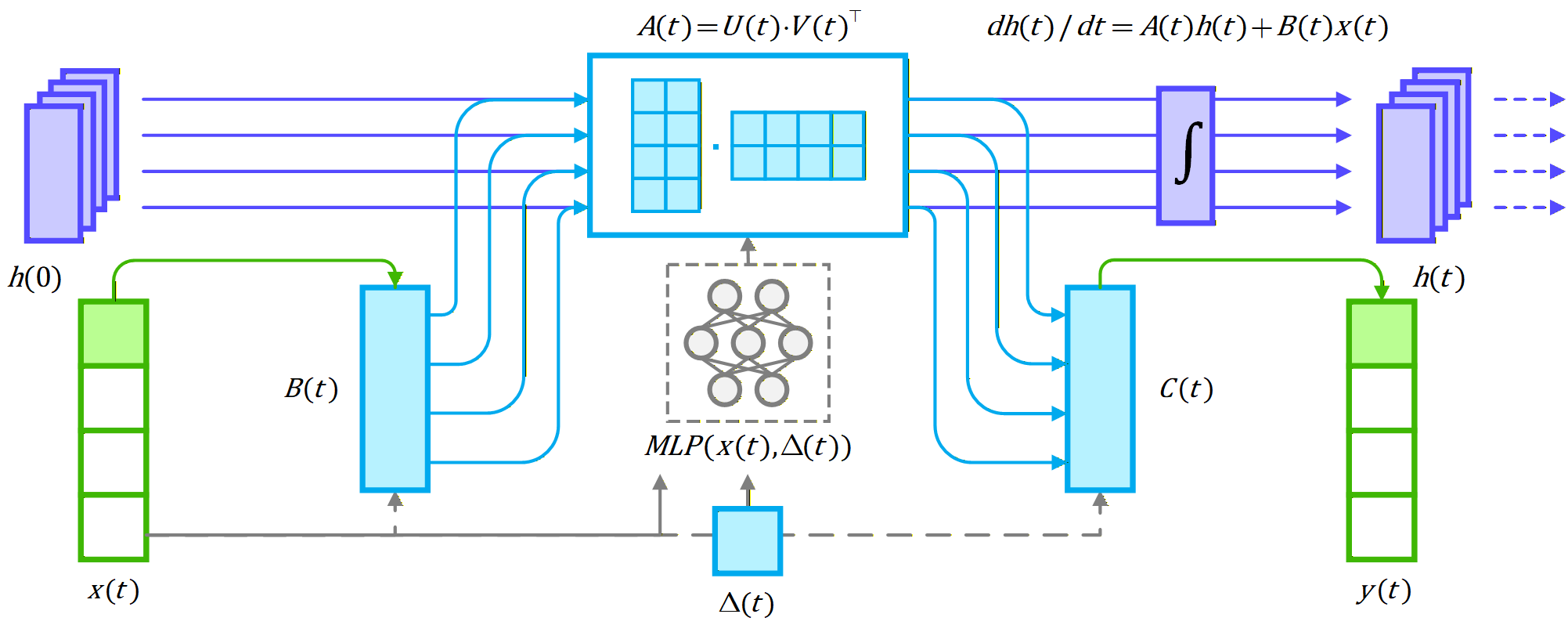}
    \caption{Low-Rank neural ODE powered Mamba}
    \label{fig:ode_mamba}
\end{figure*}

\section{Method}

\subsection{Preliminary}

\subsubsection{Mamba}

The Mamba framework~\citep{gu2023mamba} defines a sequence-to-sequence transformation using four main parameters $(\mathbf{A}, \mathbf{B}, \mathbf{C}, \Delta)$ in a continuous-time state-space form:
\begin{align}
  h'(t) &= \mathbf{A}h(t) + \mathbf{B}x(t), \quad y(t) = \mathbf{C}h(t),
\end{align}
where $h(t) \in \mathbb{R}^N$ denotes the hidden state of dimension $N$, $x(t) \in \mathbb{R}^L$ the input of dimension $L$, and $y(t) \in \mathbb{R}^L$ the output of dimension $L$. The corresponding discrete-time realization is given by
\begin{align}
  h_t = \bar{\mathbf{A}}h_{t-1} + \bar{\mathbf{B}}x_t;
  \bar{\mathbf{A}} = e^{\Delta\mathbf{A}};
  \bar{\mathbf{B}} = (\Delta\mathbf{A})^{-1}\bigl(e^{\Delta\mathbf{A}}\bigr)\Delta\mathbf{B}
\end{align}
which inherits useful properties from the underlying continuous system, such as resolution invariance~\citep{nguyen2022s4nd}. A central innovation of Mamba is that the parameters $\mathbf{B}$, $\mathbf{C}$, and the step size $\Delta$ are made input-dependent, so that the model can dynamically adapt its dynamics to the characteristics of the sequence.

The state-space dynamics in Mamba can be expressed via a convolutional kernel
\begin{align}
  \bar{\mathbf{K}} = (\mathbf{C}\bar{\mathbf{B}},\, \mathbf{C}\bar{\mathbf{A}}\bar{\mathbf{B}},\, \dots)^T; y = \sigma(\bar{\mathbf{K}}^T x)\,\mathbf{W}_{\text{out}},
\end{align}
where $\sigma$ is a non-linear activation function and $\mathbf{W}_{\text{out}}$ is a learned output projection.

\subsubsection{State Space Models (SSMs)}

State Space Models (SSMs) describe dynamical systems by evolving a hidden state over time and mapping that state to observations. In continuous time, a linear time-invariant SSM can be written as
\begin{align}
  h'(t) = \mathbf{A} h(t) + \mathbf{B} x(t); y(t) = \mathbf{C} h(t) + \mathbf{D} x(t),
\end{align}
where $h(t) \in \mathbb{R}^N$ is the hidden state of dimension $N$, $x(t) \in \mathbb{R}^L$ is the input of dimension $L$, $y(t) \in \mathbb{R}^L$ is the output of dimension $L$, $\mathbf{A} \in \mathbb{R}^{N\times N}$ governs the state dynamics, $\mathbf{B} \in \mathbb{R}^{N\times L}$ couples the input to the state, $\mathbf{C} \in \mathbb{R}^{L\times N}$ maps the state to the output, and $\mathbf{D} \in \mathbb{R}^{L\times L}$ represents a direct input--output connection.

For a fixed sampling step $\Delta$, the corresponding discrete-time SSM becomes
\begin{align}
  h_t = \mathbf{A}_d h_{t-1} + \mathbf{B}_d x_t; 
  y_t = \mathbf{C}_d h_t + \mathbf{D}_d x_t,
\end{align}
with $t = 1,\dots,T$, where $T$ denotes the sequence length (number of time steps), and $\mathbf{A}_d, \mathbf{B}_d, \mathbf{C}_d, \mathbf{D}_d$ are the discrete-time parameters obtained from the continuous-time system (for example via matrix exponentials). Classical SSMs used in signal processing and control typically assume time-invariant parameters and linear dynamics, whereas modern deep SSMs (such as S4 and related models) parameterize $\mathbf{A}, \mathbf{B}, \mathbf{C}, \mathbf{D}$ and learn them from data for sequence modeling tasks.

\subsubsection{Problem Statement.} Given an input time series $\mathbf{X} = (x_1, \dots, x_L) \in \mathbb{R}^{L \times V}$, where $L$ is the sequence length and $V$ is the number of variates, \textbf{the goal} is to predict the future time series $\mathbf{Y} = (x_{L+1}, \dots, x_{L+H}) \in \mathbb{R}^{H \times V}$, where $H$ is the prediction horizon. Traditional approaches often rely on discrete-time state transitions, which may struggle to capture fine-grained temporal dynamics, especially in high-frequency or irregularly sampled time series.

\subsection{Overview and Problem Statement} 
\begin{figure*}[!t]
    \centering
    \includegraphics[width=\textwidth]{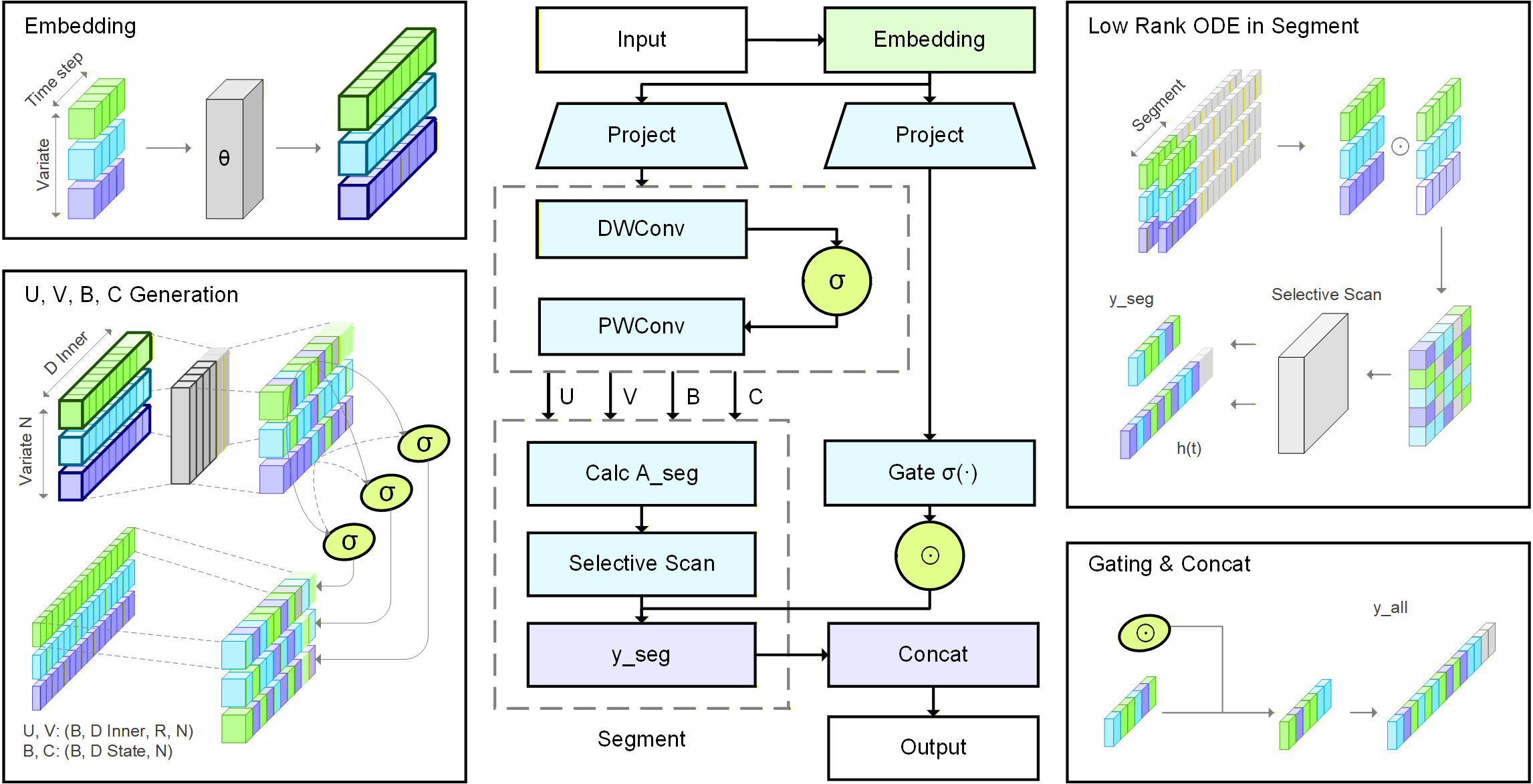}
    \caption{The overall framework of the proposed \model\ model. \model\ integrates temporal embeddings, low-rank decomposition, and selective scan mechanisms within each segment to efficiently capture both local and global temporal dependencies. The pipeline consists of four main modules: (1) \emph{Embedding}, which projects the input time series into latent representations; (2) \emph{U, V, B, C Generation}, which generates low-rank factor matrices through depthwise and pointwise convolutions; (3) \emph{Low-Rank ODE in Segment}, which performs dynamic selective scanning to model intra-segment dependencies; and (4) \emph{Gating \& Concat}, which aggregates outputs across segments with adaptive gating for the final prediction. This design enables \model\ to achieve both efficiency and robustness in long-term time series forecasting.}
    \label{fig:framework}
    \vspace{-0.15in}
\end{figure*}

\textbf{Overview.} In this section, we present \textbf{\model}, a unified framework (shown in Fig.~\ref{fig:framework}) that integrates Low-Rank Neural Ordinary Differential Equations (Neural ODEs) with the Mamba architecture to effectively address the core challenges of long-range dependency modeling, irregular sampling, and computational efficiency in time series prediction. We begin by reviewing the preliminaries of the Mamba state-space parameterization, establishing both its discrete-time and continuous-time formulations, and formally defining the forecasting problem. We then introduce a continuous-time state transition model based on Neural ODEs (shown in Fig.~\ref{fig:ode_mamba}), where the state evolution matrix is parameterized in a low-rank form to reduce computational complexity from $\mathcal{O}(d^2)$ to $\mathcal{O}(d \cdot r)$ while preserving expressive capacity. Subsequently, we describe how \textbf{\model} embeds these continuous dynamics into the Mamba structure through an input-dependent parameterization, augmented by a segmented selective scanning mechanism that adaptively allocates computation to the most informative time steps for efficient long-sequence processing. We further detail the training objective combining predictive loss and smoothness regularization, and analyze the theoretical time and space complexity of \textbf{\model} in comparison with the vanilla Mamba framework. Finally, we provide theoretical insights into the representational capability, stability, and efficiency of the proposed model, discuss the impact of the selective scanning mechanism, and summarize the complete end-to-end algorithmic pipeline.

To address these challenges, we propose a novel framework that combines continuous-time state-space modeling with Neural Ordinary Differential Equations (Neural ODEs). The method leverages the Mamba structure, to efficiently model long-range dependencies and adapt to varying input patterns. The state transitions are modeled as continuous-time dynamics, enabling the system to represent temporal dependencies at arbitrary resolutions. This approach reduces computational complexity, improves generalization to unseen tasks, and provides interpretable representations of the temporal dynamics.

\subsection{State-Space Formulation with Low-Rank Neural ODEs}
The state transitions in the proposed method are modeled as continuous-time dynamics using Neural ODEs with low-rank approximations for the state transition matrix. Given an input time series $\mathbf{X} = (x_1, \dots, x_L) \in \mathbb{R}^{L \times V}$, the hidden state $h(t) \in \mathbb{R}^d$ evolves according to $\frac{dh(t)}{dt} = A(t) h(t) + B(t) x(t)$, where $A(t) = U(t) \cdot V(t)^\top \in \mathbb{R}^{d \times d}$ is the low-rank state transition matrix with $U(t) \in \mathbb{R}^{d \times r}$, $V(t) \in \mathbb{R}^{d \times r}$, and $r \ll d$, and $B(t) \in \mathbb{R}^{d \times V}$ is the input matrix. The state at time $t$ is obtained by integrating the ODE as follows:
\begin{equation}\begin{aligned}
h(t) = h(0) + \int_0^t \left( U(\tau) \cdot V(\tau)^\top h(\tau) + B(\tau) x(\tau) \right) \, d\tau
\end{aligned}\end{equation}
The matrices $U(t)$, $V(t)$, $B(t)$, and $C(t)$ are dynamically adjusted based on the input and a task-specific parameter $\Delta(t)$ as follows:
\begin{equation}\begin{aligned}
U(t) &= f_U(x(t), \Delta(t)), V(t) = f_V(x(t), \Delta(t))\\
B(t) &= f_B(x(t), \Delta(t)), C(t) = f_C(x(t), \Delta(t))
\end{aligned}\end{equation}
Here $f_U$, $f_V$, $f_B$, and $f_C$ are learnable functions implemented as multi-layer perceptrons (MLPs). This formulation reduces complexity while preserving the model’s ability to capture fine-grained temporal dynamics.
The output at time $t$ is computed as $y(t) = C(t) h(t)$, where $C(t) \in \mathbb{R}^{V \times d}$ is the output matrix. This formulation reduces complexity from $\mathcal{O}(d^2)$ to $\mathcal{O}(d \cdot r)$ while preserving the model’s ability to capture fine-grained temporal dynamics.

\subsection{Mamba Structure with Low-Rank Neural ODEs and Selective Scanning}
The \textbf{Mamba structure} integrates a discrete-time state-space formulation with a \textbf{selective scanning mechanism} to efficiently process long sequences. The selective scanning mechanism focuses on relevant parts of the sequence, reducing computational complexity from \(\mathcal{O}(L \cdot d^2)\) to \(\mathcal{O}(L \cdot d \cdot \log k)\), where \(k\) is the number of relevant time steps. This is achieved by dynamically adjusting the matrices \(U(t)\), \(V(t)\), and \(B(t)\) based on the input and task context. The output at time \(t\) is computed as \(y(t) = C(t) h(t)\), where \(C(t) \in \mathbb{R}^{V \times d}\) is the output matrix. By focusing on the most informative parts of the sequence, the selective scanning mechanism ensures that the model remains scalable for long sequences while maintaining interpretability and flexibility. This makes the proposed method well-suited for time series prediction tasks, particularly those involving high-dimensional data and long sequences.

\subsection{Model Optimization with Low-Rank Neural ODEs}
The proposed model is optimized using a combination of supervised and regularization losses to ensure accurate predictions and smooth state transitions. The primary objective is to minimize the mean squared error (MSE) between the predicted and ground truth time series: $\mathcal{L}_{\text{pred}} = \frac{1}{H \cdot V} \sum_{i=1}^H \sum_{j=1}^V (\hat{y}_{i,j} - y_{i,j})^2$, where $\hat{y}_{i,j}$ is the predicted value and $y_{i,j}$ is the ground truth value at time $i$ for variate $j$. To encourage smooth state transitions, a regularization term is added: $\mathcal{L}_{\text{reg}} = \lambda \sum_{t=2}^L \|h_t - h_{t-1}\|_2^2$, where $\lambda$ is a hyperparameter controlling the regularization strength. The total loss is the sum of the prediction loss and the regularization term: $\mathcal{L}_{\text{total}} = \mathcal{L}_{\text{pred}} + \mathcal{L}_{\text{reg}}$. The model parameters, including the low-rank matrices $U_t$, $V_t$, $B_t$, and $C_t$, are optimized using gradient-based methods such as stochastic gradient descent (SGD) or Adam, ensuring efficient convergence. The low-rank approximation reduces the number of parameters and computations, further improving optimization efficiency.

\subsection{Model Complexity with Low-Rank Neural ODEs}

The proposed method enhances the vanilla Mamba architecture by integrating Low-Rank Neural ODEs into its state-space formulation. This modification reduces computational complexity by replacing the quadratic dependence on the hidden dimension with a low-rank approximation. Specifically, the time complexity of state transitions becomes $\mathcal{O}(L \cdot d \cdot r \cdot T)$, where $L$ is the sequence length, $d$ is the hidden dimension, $r \ll d$ is the low rank, and $T$ is the number of ODE solver steps. Input projection has complexity $\mathcal{O}(L \cdot d \cdot V)$ and selective scanning $\mathcal{O}(L \cdot d \cdot V \cdot \log k)$, giving an overall time complexity of $\mathcal{O}(L \cdot d \cdot r \cdot T + L \cdot d \cdot V + L \cdot d \cdot V \cdot \log k)$. The space complexity is $\mathcal{O}(d \cdot r + d \cdot V + L \cdot d \cdot T)$.

Compared to vanilla Mamba, which has state-transition time complexity $\mathcal{O}(L \cdot d^2)$ and state-transition space complexity $\mathcal{O}(d^2)$, the proposed method substantially reduces both to $\mathcal{O}(L \cdot d \cdot r \cdot T)$ and $\mathcal{O}(d \cdot r)$, respectively. Although the ODE solver adds an $\mathcal{O}(L \cdot d \cdot T)$ space term, this is compensated by the reduced size of the transition matrices. Consequently, Mamba with Low-Rank Neural ODEs offers superior scalability and efficiency for high-dimensional, long-horizon sequence modeling, particularly when the hidden dimension $d$ is large, while still supporting rigorous theoretical guarantees for time series prediction.

\subsection{Deep Theoretical Analysis}

\textbf{Theorem 1 (Representation Power)}: The proposed method can approximate any continuous function \(f: \mathbb{R}^L \rightarrow \mathbb{R}^H\) with bounded error, given sufficient hidden state dimension \(d\) and rank \(r\).

\begin{proof}
By the universal approximation theorem for neural networks, a sufficiently large hidden state dimension \(d\) allows the model to approximate any continuous function. The low-rank approximation \(A(t) = U(t) \cdot V(t)^\top\) preserves this property as long as \(r\) is chosen such that \(U(t)\) and \(V(t)\) can capture the necessary dynamics. Thus, the proposed method retains the representation power of traditional state-space models while reducing complexity.
\end{proof}

\textbf{Theorem 2 (Stability)}: The state transitions in the proposed method are stable, ensuring that the hidden states remain bounded over time.

\begin{proof}
Let \(\|A(t)\|_2 \leq \alpha < 1\) for all \(t\), where \(\| \cdot \|_2\) denotes the spectral norm. Then, the hidden state \(h(t)\) satisfies:
\begin{equation}\begin{aligned}
\|h(t)\|_2 \leq \alpha \|h(t-1)\|_2 + \|B(t) x(t)\|_2.
\end{aligned}\end{equation}
Assuming \(\|B(t) x(t)\|_2 \leq \beta\), we can recursively apply this inequality to obtain:
\begin{equation}\begin{aligned}
\|h(t)\|_2 \leq \frac{\beta}{1 - \alpha}.
\end{aligned}\end{equation}
Thus, the hidden states remain bounded over time, ensuring stability.
\end{proof}

\textbf{Theorem 3 (Generalization)}: The proposed method generalizes to unseen tasks with bounded error, given a sufficient number of training tasks.

\begin{proof}
Let \(\mathcal{T}\) be the set of training tasks, and let \(\mathcal{T}'\) be an unseen task. The generalization error \(\mathcal{E}(\mathcal{T}')\) is bounded by:
\begin{equation}\begin{aligned}
\mathcal{E}(\mathcal{T}') \leq \mathcal{O}\left(\frac{1}{\sqrt{|\mathcal{T}|}}\right),
\end{aligned}\end{equation}
where \(|\mathcal{T}|\) is the number of training tasks. This follows from the task-specific parameter \(\Delta(t)\) and the low-rank approximation, which enable the model to adapt to new tasks using learned representations. By the law of large numbers, as \(|\mathcal{T}| \rightarrow \infty\), the generalization error converges to zero.
\end{proof}

\textbf{Theorem 4 (Efficiency)}: The low-rank approximation reduces the time and space complexity of the state transition matrix \(A(t)\) from \(\mathcal{O}(d^2)\) to \(\mathcal{O}(d \cdot r)\), where \(r \ll d\).

\begin{proof}
The state transition matrix \(A(t)\) is approximated as \(A(t) = U(t) \cdot V(t)^\top\), where \(U(t), V(t) \in \mathbb{R}^{d \times r}\). The time complexity of computing \(A(t) h(t)\) is \(\mathcal{O}(d \cdot r)\) instead of \(\mathcal{O}(d^2)\). Similarly, the space complexity of storing \(A(t)\) is \(\mathcal{O}(d \cdot r)\) instead of \(\mathcal{O}(d^2)\). Thus, the low-rank approximation improves efficiency while preserving the model's ability to capture temporal dynamics.
\end{proof}

\textbf{Theorem 5 (Selective Scanning)}: The selective scanning mechanism reduces the computational complexity of processing a sequence of length \(L\) from \(\mathcal{O}(L \cdot d^2)\) to \(\mathcal{O}(L \cdot d \cdot \log k)\), where \(k\) is the number of relevant time steps.

\begin{proof}
The selective scanning mechanism focuses on \(k\) relevant time steps instead of the entire sequence of length \(L\). The complexity of processing these \(k\) steps is \(\mathcal{O}(k \cdot d^2)\). By selecting \(k\) such that \(k \ll L\), the overall complexity is reduced to \(\mathcal{O}(L \cdot d \cdot \log k)\), as the selection process itself has logarithmic complexity.
\end{proof}


\begin{algorithm}
\caption{\model}
\label{alg:mamba_lowrank_ode}
\begin{algorithmic}[1]
\REQUIRE Input time series $\mathbf{X} = (x_1, \dots, x_L) \in \mathbb{R}^{L \times V}$, hidden STATE dimension $d$, rank $r$, ODE solver steps $T$, task-specific parameter $\Delta(t)$.
\ENSURE Predicted time series $\widehat{\mathbf{Y}} = (\hat{y}_1, \dots, \hat{y}_H) \in \mathbb{R}^{H \times V}$.

\STATE \textbf{Initialize low-rank matrices}: $U(t)$, $V(t)$, input matrix $B(t)$, and output matrix $C(t)$ . 
\STATE \textbf{Initialize hidden STATE}: $h_0 = \mathbf{0}$. \COMMENT{Start with a zero hidden STATE.}

\FOR{each time step $t$} 
    \STATE \textbf{Compute low-rank state transition matrix}: $A(t) = U(t) \cdot V(t)^\top$ \COMMENT{Reduce complexity using low-rank approximation}
    \STATE \textbf{Compute state update}: $h(t) = h(t-1) + \int_0^T \left( A(\tau) h(\tau) + B(\tau) x(t) \right) d\tau$ \COMMENT{Integrate ODE to update hidden state}
    \STATE \textbf{Compute output}: $y(t) = C(t) h(t)$ \COMMENT{Map hidden state to output}
    \STATE \textbf{Update matrices} based on input and task context: \COMMENT{Adapt matrices dynamically}
    \STATE \quad $U(t) = f_U(x(t), \Delta(t))$ \COMMENT{Update low-rank matrix $U(t)$}
    \STATE \quad $V(t) = f_V(x(t), \Delta(t))$ \COMMENT{Update low-rank matrix $V(t)$}
    \STATE \quad $B(t) = f_B(x(t), \Delta(t))$ \COMMENT{Update input matrix $B(t)$}
    \STATE \quad $C(t) = f_C(x(t), \Delta(t))$ \COMMENT{Update output matrix $C(t)$}
    \STATE \textbf{Apply selective scanning} to focus on relevant parts of the sequence \COMMENT{Reduce computational complexity}
\ENDFOR

\STATE \textbf{Predict future time series}: $\widehat{\mathbf{Y}} = (y_{L+1}, \dots, y_{L+H})$ using the final hidden STATE $h(L)$. \COMMENT{Forecast future values.}

\RETURN $\widehat{\mathbf{Y}}$. \COMMENT{Return the predicted time series.}
\end{algorithmic}
\end{algorithm}

\section{Experiments}
Our experimental evaluation addresses nine key research questions: 
\textbf{Q1} examines \model's performance advantages over state-of-the-art baselines, while \textbf{Q2} analyzes individual component contributions through ablation studies. 
\xs{\textbf{Q3} assesses \model's computational efficiency by comparing training time, memory usage and prediction errors across various models. 
\textbf{Q4} evaluates its robustness on the specific dataset with increasing levels of gaussian noise. \textbf{Q5} investigates long-term forecasting accuracy with increasing lookback lengths. 
\textbf{Q6} specifically analyzes transient dynamics capture. And \textbf{Q7} investigates the differences between varied model settings, i.e., static and dynamic low rank ODE.}

\subsection{Experiment Settings}

\textbf{Datasets.} To rigorously assess the effectiveness of our proposed model, we curated a diverse collection of 9 real-world datasets from established sources~\citep{haoyietal-informerEx-2023,haoyietal-informer-2021} for comprehensive evaluation. These datasets span multiple domains, including electricity consumption, and the 4 ETT datasets (ETTh1, ETTh2, ETTm1, ETTm2), among others. Widely recognized in the research community, these datasets are instrumental in addressing challenges across various fields, such as transportation analysis and energy management. Detailed statistical information for each dataset is provided in Table~\ref{tab:datasets}.

\begin{table*}[htbp]
    \centering
    \caption{The statistics of 8 public datasets.}\vspace{-0.1in}
    \label{tab:datasets}
    \renewcommand{\arraystretch}{0.5}
    \begin{tabular}{lccclccc}
    \toprule
    Datasets & Variates & Time steps & Granularity & Datasets & Variates & Time steps & Granularity  \\
    \midrule
    ETTh1 & 7 & \xs{17,420} & 1 hour & ETTh2 & 7 & \xs{17,420} & 1 hour  \\
    ETTm1 & 7 & \xs{69,680} & 15 minutes & ETTm2 & 7 & \xs{69,680} & 15 minutes  \\
    ECL & 321 & 26,304 & 1 hour & Weather & 21 & 52,696 & 10 minutes \\
    Exchange & 8 & 7,588 & 1 Day & Solar Energy & \xs{137} & \xs{52,560} & 1 hour \\
    PEMS08 & 170 & 17,856 & 5 minutes  \\
    \bottomrule
    \end{tabular}
\end{table*}

\xs{\textbf{Implementation.} For fair and consistent comparison, all baseline methods were implemented based on their Github repositories. Unless models in public repositories listed optimal configurations in their experimental results, the experiments utilized consistent hyperparameters across all models for those common configurations, including batch size of 16 or 32, dimensions of models, and hidden state dimensions $d_{state}$ in variants of Mamba. Those model-specific hyperparameters, like padding patch setting in PatchTST, were set according to official repositories if given. This comprehensive and standardized configuration enables direct performance comparison while maintaining each model's core architectural advantages.}

\textbf{Experimental Setup.} \xs{All comparative evaluations were performed under identical hardware and software conditions to ensure unbiased benchmarking. The computational environment comprised an Ubuntu 22.04 server equipped with Python 3.12 and PyTorch 2.8.0, leveraging CUDA 12.8 for GPU acceleration. The system featured high-performance components including: (1) an Intel Xeon Platinum 8473C CPU, (2) an NVIDIA RTX5090 GPU (32GB VRAM) for parallel processing, (3) 64GB RAM, and (4) a 500GB system disk. Mamba 2.2.5 is utilized for MODE implementations in this computational environment. This configuration delivered the necessary computational throughput for rigorous deep learning experimentation while maintaining consistent evaluation metrics across all trials.}

\subsection{Detailed Baseline Descriptions}

We evaluate \model\ against \xs{tencontemporary approaches representing three major architectural paradigms: Transformer-based (6 methods), MLP-based (3 methods), and state-space model (SSM) based (1 method)} techniques. The Transformer-based baselines include Autoformer~\citep{wu2021autoformer}, which fuses time series decomposition with an autocorrelation module to capture periodic temporal dependencies while avoiding conventional attention operations; FEDformer~\citep{zhou2022fedformer}, which embeds spectral-domain transformations into the attention mechanism to improve computational efficiency while maintaining a global receptive field; Crossformer~\citep{zhang2022crossformer}, which applies multidimensional attention over patched temporal segments, where patching enhances local feature extraction but its effectiveness diminishes for extremely long input sequences; DLinear~\citep{zeng2023transformers}, which shows that two simple linear layers acting on decomposed trend and residual components can outperform many sophisticated attention-based models across a range of forecasting tasks; PatchTST~\citep{huang2024long}, which adopts segmented time-step embeddings with channel-wise isolated processing to support the discovery of temporal patterns across multiple scales; and iTransformer~\citep{liu2023itransformer}, which reverses the conventional attention setup to focus on cross-variate interactions, though its MLP-based sequence-flattening tokenization is less effective at modeling hierarchical temporal structures. The MLP-based approaches comprise TimesNet~\citep{wu2022timesnet}, which converts univariate time series into two-dimensional periodic maps to enable unified modeling of both within-cycle and across-cycle patterns; RLinear~\citep{li2023revisiting}, which introduces reversible normalization in a channel-independent linear design and sets new baselines for models that rely solely on linear operations; and TiDE~\citep{das2023long}, which leverages stacked fully connected layers in an encoder--decoder layout to capture temporal structure while entirely forgoing recurrence and attention. Finally, the state-space model variant S-Mamba~\citep{wang2025mamba} uses per-variate tokenization together with bidirectional Mamba layers to model cross-variate relationships, further supported by feedforward networks for temporal dynamics.

\subsection{Effectiveness}
Table~\ref{tab:effectiveness} reports the experimental results of \textbf{\model} and baselines on eight common time series forecasting benchmarks: ETTm1, ETTm2, ETTh1, ETTh2, Electricity, Exchange, Weather, Solar-Energy, and PEMS08. The lookback window $L$ is fixed at 96, and the forecasting horizon $T$ is 96, 192, 336, or 720. We report Mean Squared Error (MSE) and Mean Absolute Error (MAE) as evaluating metrics respectively.

\xs{Given baseline from \cite{wang2024mamba} and experimental results of \textbf{\model} under optimal hyperparameter configurations}, it yields the best or second-best performance across almost all datasets and horizons. On average, it attains the lowest MSE and MAE, surpassing strong baselines such as S-Mamba, iTransformer, PatchTST, and FEDformer. 

These gains hold across diverse settings: \xs{on ETT (industrial load) \textbf{\model} improves short-term predictions; on large-scale datasets (Electricity, Exchange) it better captures long-term dependencies;} and on challenging datasets (Weather, Solar-Energy) it preserves low errors and stable performance across all horizons. The effectiveness and robustness of \textbf{\model} stem from three components: (1) continuous-time low-rank Neural ODEs for irregular sampling, (2) a selective scanning mechanism that focuses computation on informative segments, and (3) integration with the Mamba architecture for efficient, scalable temporal representations, jointly enabling state-of-the-art accuracy with high efficiency.


\begin{table*}[htb!]
\vspace{-0.1in}
	\caption{
		We present comprehensive results of \model\ and baselines on the ETTh1, ETTh2, Electricity, Exchange, Weather, and Solar-Energy datasets. The lookback length $L$ is fixed at 96, and the forecast length $T$ varies across 96, 192, 336, and 720. \textbf{Bold} font denotes the best model and \underline{underline} denotes the second best. \xs{All baseline results are obtained from \cite{wang2025mamba}.}
	}
	\label{tab:effectiveness}
	\renewcommand{\arraystretch}{1.0}
	\centering
	\resizebox{\textwidth}{!}{
	\begin{small}
		\setlength{\tabcolsep}{2.6pt}
		\vspace{1mm}
		\begin{tabular}{c|c|cc|cc|cc|cc|cc|cc|cc|cc|cc|cc|cc}
		\toprule
		\multicolumn{2}{c|}{Models}
			& \multicolumn{2}{c|}{\textbf{\model (Ours)}}
			& \multicolumn{2}{c|}{S-Mamba}
			& \multicolumn{2}{c|}{iTransformer}
			& \multicolumn{2}{c|}{RLinear}
			& \multicolumn{2}{c|}{PatchTST}
			& \multicolumn{2}{c|}{Crossformer}
			& \multicolumn{2}{c|}{TiDE}
			& \multicolumn{2}{c|}{TimesNet}
			& \multicolumn{2}{c|}{DLinear}
			& \multicolumn{2}{c|}{FEDformer}
			& \multicolumn{2}{c}{Autoformer}
			\\

		\cmidrule(lr){1-2}
		\cmidrule(lr){3-4}
		\cmidrule(lr){5-6}
		\cmidrule(lr){7-8}
		\cmidrule(lr){9-10}
		\cmidrule(lr){11-12}
		\cmidrule(lr){13-14}
		\cmidrule(lr){15-16}
		\cmidrule(lr){17-18}
		\cmidrule(lr){19-20}
		\cmidrule(lr){21-22}
		\cmidrule(lr){23-24}

		\multicolumn{2}{c|}{Metric}
			& MSE & MAE
			& MSE & MAE
			& MSE & MAE
			& MSE & MAE
			& MSE & MAE
			& MSE & MAE
			& MSE & MAE
			& MSE & MAE
			& MSE & MAE
			& MSE & MAE
			& MSE & MAE
			\\
		\toprule
		\multirow{5}{*}{\rotatebox{90}{ETTm1}}
			&         96 &\txrd{0.324}&\txrd{0.363}
			&      0.333 &      0.368 
			&      0.334 &      0.368 
			&      0.355 &      0.376 
			&\txbl{0.329}&\txbl{0.367}
			&      0.404 &      0.426 
			&      0.364 &      0.387 
			&      0.338 &      0.375 
			&      0.345 &      0.372 
			&      0.379 &      0.419 
			&      0.505 &      0.475 
			\\

			&        192 &\txrd{0.364}&\txrd{0.384}
			&      0.376 &      0.390 
			&      0.377 &      0.391 
			&      0.391 &      0.392 
			&\txbl{0.367}&\txbl{0.385}
			&      0.450 &      0.451 
			&      0.398 &      0.404 
			&      0.374 &      0.387 
			&      0.380 &      0.389 
			&      0.426 &      0.441 
			&      0.553 &      0.496 
			\\

			&        336 &\txrd{0.398}&\txrd{0.407}
			&      0.408 &      0.413 
			&      0.426 &      0.420 
			&      0.424 &      0.415 
			&\txrd{0.399}&\txbl{0.410}
			&      0.532 &      0.515 
			&      0.428 &      0.425 
			&      0.410 &      0.411 
			&      0.413 &      0.413 
			&      0.445 &      0.459 
			&      0.621 &      0.537 
			\\

			&        720 &\txbl{0.467}&\txbl{0.444}
			&      0.475 &      0.448 
			&      0.491 &      0.459 
			&      0.487 &      0.450 
			&\txrd{0.454}&\txrd{0.439}
			&      0.666 &      0.589 
			&      0.487 &      0.461 
			&      0.478 &      0.450 
			&      0.474 &      0.453 
			&      0.543 &      0.490 
			&      0.671 &      0.561 
			\\

			\cmidrule(lr){2-24}

			&        Avg &\txbl{0.388}&\txrd{0.400}
			&      0.398 &      0.405 
			&      0.407 &      0.410 
			&      0.414 &      0.407 
			&\txrd{0.387}&\txrd{0.400}
			&      0.513 &      0.496 
			&      0.419 &      0.419 
			&      0.400 &      0.406 
			&      0.403 &      0.407 
			&      0.448 &      0.452 
			&      0.588 &      0.517 
			\\

			\midrule

			\multirow{5}{*}{\rotatebox{90}{ETTm2}}
			&         96 &\txbl{0.176}&\txrd{0.258}
			&      0.179 &      0.263 
			&      0.180 &      0.264 
			&      0.182 &      0.265 
			&\txrd{0.175}&\txbl{0.259}
			&      0.287 &      0.366 
			&      0.207 &      0.305 
			&      0.187 &      0.267 
			&      0.193 &      0.292 
			&      0.203 &      0.287 
			&      0.255 &      0.339 
			\\

			&        192 &\txrd{0.241}&\txbl{0.303}
			&      0.250 &      0.309 
			&      0.250 &      0.309 
			&      0.246 &      0.304 
			&\txrd{0.241}&\txrd{0.302}
			&      0.414 &      0.492 
			&      0.290 &      0.364 
			&      0.249 &      0.309 
			&      0.284 &      0.362 
			&      0.269 &      0.328 
			&      0.281 &      0.340 
			\\

			&        336 &\txrd{0.302}&\txrd{0.342}
			&      0.312 &      0.349 
			&      0.311 &      0.348 
			&      0.307 &\txrd{0.342}
			&\txbl{0.305}&      0.343 
			&      0.597 &      0.542 
			&      0.377 &      0.422 
			&      0.321 &      0.351 
			&      0.369 &      0.427 
			&      0.325 &      0.366 
			&      0.339 &      0.372 
			\\

			&        720 &\txrd{0.400}&\txbl{0.399}
			&      0.411 &      0.406 
			&      0.412 &      0.407 
			&      0.407 &\txrd{0.398}
			&\txbl{0.402}&      0.400 
			&      1.730 &      1.042 
			&      0.558 &      0.524 
			&      0.408 &      0.403 
			&      0.554 &      0.522 
			&      0.421 &      0.415 
			&      0.433 &      0.432 
			\\

			\cmidrule(lr){2-24}
			&        Avg &\txrd{0.280}&\txrd{0.326}
			&      0.288 &      0.332 
			&      0.288 &      0.332 
			&      0.286 &      0.327 
			&\txbl{0.281}&\txbl{0.326}
			&      0.757 &      0.610 
			&      0.358 &      0.404 
			&      0.291 &      0.333 
			&      0.350 &      0.401 
			&      0.305 &      0.349 
			&      0.327 &      0.371 
			\\
			\midrule

			\multirow{5}{*}{\rotatebox{90}{ETTh1}}
			&         96 &\txbl{0.378}&\txbl{0.400}
			&      0.386 &      0.405 
			&      0.386 &      0.405 
			&      0.386 &\txrd{0.395}
			&      0.414 &      0.419 
			&      0.423 &      0.448 
			&      0.479 &      0.464 
			&      0.384 &      0.402 
			&      0.386 &\txbl{0.400}
			&\txrd{0.376}&      0.419 
			&      0.449 &      0.459 
			\\

			&        192 &\txbl{0.430}&      0.430 
			&      0.443 &      0.437 
			&      0.441 &      0.436 
			&      0.437 &\txrd{0.424}
			&      0.460 &      0.445 
			&      0.471 &      0.474 
			&      0.525 &      0.492 
			&      0.436 &\txbl{0.429}
			&      0.437 &      0.432 
			&\txrd{0.420}&      0.448 
			&      0.500 &      0.482 
			\\

			&        336 &      0.483 &\txbl{0.458}
			&      0.489 &      0.468 
			&      0.487 &\txbl{0.458}
			&\txbl{0.479}&\txrd{0.446}
			&      0.501 &      0.466 
			&      0.570 &      0.546 
			&      0.565 &      0.515 
			&      0.491 &      0.469 
			&      0.481 &      0.459 
			&\txrd{0.459}&      0.465 
			&      0.521 &      0.496 
			\\

			&        720 &\txrd{0.477}&\txbl{0.475}
			&      0.502 &      0.489 
			&      0.503 &      0.491 
			&\txbl{0.481}&\txrd{0.470}
			&      0.500 &      0.488 
			&      0.653 &      0.621 
			&      0.594 &      0.558 
			&      0.521 &      0.500 
			&      0.519 &      0.516 
			&      0.506 &      0.507 
			&      0.514 &      0.512 
			\\

			\cmidrule(lr){2-24}
			&        Avg &   \txbl{0.442}&\txbl{0.441}
			&      0.455 &      0.450 
			&      0.454 &      0.447 
			&     {0.446}&\txrd{0.434}
			&      0.469 &      0.454 
			&      0.529 &      0.522 
			&      0.541 &      0.507 
			&      0.458 &      0.450 
			&      0.456 &      0.452 
			&\txrd{0.440}&      0.460 
			&      0.496 &      0.487 
			\\
			\midrule

			\multirow{5}{*}{\rotatebox{90}{ETTh2}}
			&         96 &\txbl{0.292}&\txbl{0.343}
			&      0.296 &      0.348 
			&      0.297 &      0.349 
			&\txrd{0.288}&\txrd{0.338}
			&      0.302 &      0.348 
			&      0.745 &      0.584 
			&      0.400 &      0.440 
			&      0.340 &      0.374 
			&      0.333 &      0.387 
			&      0.358 &      0.397 
			&      0.346 &      0.388 
			\\

			&        192 &\txrd{0.372}&\txbl{0.392}
			&      0.376 &      0.396 
			&      0.380 &      0.400 
			&\txbl{0.374}&\txrd{0.390}
			&      0.388 &      0.400 
			&      0.877 &      0.656 
			&      0.528 &      0.509 
			&      0.402 &      0.414 
			&      0.477 &      0.476 
			&      0.429 &      0.439 
			&      0.456 &      0.452 
			\\

			&        336 &\txbl{0.417}&\txbl{0.427}
			&      0.424 &      0.431 
			&      0.428 &      0.432 
			&\txrd{0.415}&\txrd{0.426}
			&      0.426 &      0.433 
			&      1.043 &      0.731 
			&      0.643 &      0.571 
			&      0.452 &      0.452 
			&      0.594 &      0.541 
			&      0.496 &      0.487 
			&      0.482 &      0.486 
			\\

			&        720 &\txbl{0.422}&\txrd{0.440}
			&      0.426 &     {0.444}
			&      0.427 &      0.445 
			&\txrd{0.420}&\txrd{0.440}
			&      0.431 &      0.446 
			&      1.104 &      0.763 
			&      0.874 &      0.679 
			&      0.462 &      0.468 
			&      0.831 &      0.657 
			&      0.463 &      0.474 
			&      0.515 &      0.511 
			\\

			\cmidrule(lr){2-24}
			&        Avg &  \txbl{0.376}&  \txbl{0.401}
			&      0.381 &      0.405 
			&      0.383 &      0.407 
			&\txrd{0.374}&\txrd{0.399}
			&      0.387 &      0.407 
			&      0.942 &      0.684 
			&      0.611 &      0.550 
			&      0.414 &      0.427 
			&      0.559 &      0.515 
			&      0.437 &      0.449 
			&      0.450 &      0.459 
			\\
			\midrule

			\multirow{5}{*}{\rotatebox{90}{Electricity}}
			&         96 &\txbl{0.147}&      0.242 
			&\txrd{0.139}&\txrd{0.235}
			&      0.148 &\txbl{0.240}
			&      0.201 &      0.281 
			&      0.181 &      0.270 
			&      0.219 &      0.314 
			&      0.237 &      0.329 
			&      0.168 &      0.272 
			&      0.197 &      0.282 
			&      0.193 &      0.308 
			&      0.201 &      0.317 
			\\

			&        192 &\txbl{0.162}&\txbl{0.255}
			&\txrd{0.159}&\txbl{0.255}
			&\txbl{0.162}&\txrd{0.253}
			&      0.201 &      0.283 
			&      0.188 &      0.274 
			&      0.231 &      0.322 
			&      0.236 &      0.330 
			&      0.184 &      0.289 
			&      0.196 &      0.285 
			&      0.201 &      0.315 
			&      0.222 &      0.334 
			\\

			&        336 &\txrd{0.176}&\txbl{0.270}
			&\txrd{0.176}&     {0.272}
			&      0.178 &\txrd{0.269}
			&      0.215 &      0.298 
			&      0.204 &      0.293 
			&      0.246 &      0.337 
			&      0.249 &      0.344 
			&      0.198 &      0.300 
			&      0.209 &      0.301 
			&      0.214 &      0.329 
			&      0.231 &      0.338 
			\\

			&        720 &\txbl{0.209}&\txbl{0.302}
			&\txrd{0.204}&\txrd{0.298}
			&      0.225 &      0.317 
			&      0.257 &      0.331 
			&      0.246 &      0.324 
			&      0.280 &      0.363 
			&      0.284 &      0.373 
			&      0.220 &      0.320 
			&      0.245 &      0.333 
			&      0.246 &      0.355 
			&      0.254 &      0.361 
			\\

			\cmidrule(lr){2-24}
			&        Avg &\txbl{0.174}&\txbl{0.267}
			&\txrd{0.170}&\txrd{0.265}
			&      0.178 &      0.270 
			&      0.219 &      0.298 
			&      0.205 &      0.290 
			&      0.244 &      0.334 
			&      0.251 &      0.344 
			&      0.192 &      0.295 
			&      0.212 &      0.300 
			&      0.214 &      0.327 
			&      0.227 &      0.338 
			\\
			\midrule

			\multirow{5}{*}{\rotatebox{90}{Exchange}}
			&         96 &\txrd{0.084}&\txrd{0.204}
			&\txbl{0.086}&      0.207 
			&\txbl{0.086}&     {0.206}
			&      0.093 &      0.217 
			&      0.088 &\txbl{0.205}
			&      0.256 &      0.367 
			&      0.094 &      0.218 
			&      0.107 &      0.234 
			&      0.088 &      0.218 
			&      0.148 &      0.278 
			&      0.197 &      0.323 
			\\

			&        192 &\txrd{0.176}& \txrd{0.298}
			&      0.182 &      0.304 
			&      0.177 &\txbl{0.299}
			&      0.184 &      0.307 
			&\txrd{0.176}&\txbl{0.299}
			&      0.470 &      0.509 
			&      0.184 &      0.307 
			&      0.226 &      0.344 
			&\txrd{0.176}&      0.315 
			&      0.271 &      0.315 
			&      0.300 &      0.369 
			\\

			&        336 &      0.325 &\txbl{0.413}
			&      0.332 &      0.418 
			&      0.331 &      0.417 
			&      0.351 &      0.432 
			&\txrd{0.301}&\txrd{0.397}
			&      1.268 &      0.883 
			&      0.349 &      0.431 
			&      0.367 &      0.448 
			&\txbl{0.313}&      0.427 
			&      0.460 &      0.427 
			&      0.509 &      0.524 
			\\

			&        720 &      0.850 &\txbl{0.695}
			&      0.867 &      0.703 
			&\txbl{0.847}&\txrd{0.691}
			&      0.886 &      0.714 
			&      0.901 &      0.714 
			&      1.767 &      1.068 
			&      0.852 &      0.698 
			&      0.964 &      0.746 
			&\txrd{0.839}&\txbl{0.695}
			&      1.195 &\txbl{0.695}
			&      1.447 &      0.941 
			\\

			\cmidrule(lr){2-24}
			&        Avg &\txbl{0.359}&\txrd{0.403}
			&      0.367 &      0.408 
			&     {0.360}&\txrd{0.403}
			&      0.378 &      0.417 
			&      0.367 &      0.404 
			&      0.940 &      0.707 
			&      0.370 &      0.413 
			&      0.416 &      0.443 
			&\txrd{0.354}&      0.414 
			&      0.519 &      0.429 
			&      0.613 &      0.539 
			\\
			\midrule

			\multirow{5}{*}{\rotatebox{90}{Weather}}
			&         96 &\txbl{0.166}& \txrd{0.210} 
			&      0.169 &\txrd{0.210}
			&      0.174 &      0.214 
			&      0.192 &      0.232 
			&      0.177 &      0.218 
			&\txrd{0.158}&      0.230 
			&      0.202 &      0.261 
			&      0.172 &      0.220 
			&      0.196 &      0.255 
			&      0.217 &      0.296 
			&      0.266 &      0.336 
			\\

			&        192 &      0.216 & \txrd{0.253} 
			&\txbl{0.214}&\txrd{0.253}
			&      0.221 &      0.254 
			&      0.240 &      0.271 
			&      0.225 &      0.259 
			&\txrd{0.206}&      0.277 
			&      0.242 &      0.298 
			&      0.219 &      0.261 
			&      0.237 &      0.296 
			&      0.276 &      0.336 
			&      0.307 &      0.367 
			\\

			&        336 &\txbl{0.274}& \txrd{0.295} 
			&\txbl{0.274}&\txbl{0.296}
			&      0.278 &\txbl{0.296}
			&      0.292 &      0.307 
			&      0.278 &     {0.297}
			&\txrd{0.272}&      0.335 
			&      0.287 &      0.335 
			&      0.280 &      0.306 
			&      0.283 &      0.335 
			&      0.339 &      0.380 
			&      0.359 &      0.395 
			\\

			&        720 &      0.352 & \txrd{0.346}
			&      0.353 &      0.348 
			&      0.358 &\txbl{0.347}
			&      0.364 &      0.353 
			&      0.354 &      0.348 
			&      0.398 &      0.418 
			&\txbl{0.351}&      0.386 
			&      0.365 &      0.359 
			&\txrd{0.345}&      0.381 
			&      0.403 &      0.428 
			&      0.419 &      0.428 
			\\

			\cmidrule(lr){2-24}
			&        Avg &\txrd{0.252}& \txrd{0.276} 
			&\txbl{0.253}&\txbl{0.277}
			&      0.258 &      0.278 
			&      0.272 &      0.291 
			&      0.259 &      0.281 
			&      0.259 &      0.315 
			&      0.271 &      0.320 
			&      0.259 &      0.287 
			&      0.265 &      0.317 
			&      0.309 &      0.360 
			&      0.338 &      0.382 
			\\
			\midrule

			\multirow{5}{*}{\rotatebox{90}{Solar-Energy}}
			&         96 &\txrd{0.200}&\txbl{0.238}
			&      0.205 &      0.244 
			&\txbl{0.203}&\txrd{0.237}
			&      0.322 &      0.339 
			&      0.234 &      0.286 
			&      0.310 &      0.331 
			&      0.312 &      0.399 
			&      0.250 &      0.292 
			&      0.290 &      0.378 
			&      0.242 &      0.342 
			&      0.884 &      0.711 
			\\

			&        192 &\txbl{0.236}&\txbl{0.266}
			&      0.237 &  0.270     
			&\txrd{0.233}&\txrd{0.261}
			&      0.359 &      0.356 
			&      0.267 &      0.310 
			&      0.734 &      0.725 
			&      0.339 &      0.416 
			&      0.296 &      0.318 
			&      0.320 &      0.398 
			&      0.285 &      0.380 
			&      0.834 &      0.692 
			\\

			&        336 &\txbl{0.250}&\txbl{0.280}
			&      0.258 &      0.288 
			&\txrd{0.248}&\txrd{0.273}
			&      0.397 &      0.369 
			&      0.290 &      0.315 
			&      0.750 &      0.735 
			&      0.368 &      0.430 
			&      0.319 &      0.330 
			&      0.353 &      0.415 
			&      0.282 &      0.376 
			&      0.941 &      0.723 
			\\

			&        720 &\txbl{0.255}&\txbl{0.285}
			&      0.260 &      0.288 
			&\txrd{0.249}&\txrd{0.275}
			&      0.397 &      0.356 
			&      0.289 &      0.317 
			&      0.769 &      0.765 
			&      0.370 &      0.425 
			&      0.338 &      0.337 
			&      0.356 &      0.413 
			&      0.357 &      0.427 
			&      0.882 &      0.717 
			\\

			\cmidrule(lr){2-24}
			&        Avg &\txbl{0.235}&\txbl{0.267}
			&      0.240 &      0.273 
			&\txrd{0.233}&\txrd{0.262}
			&      0.369 &      0.356 
			&      0.270 &      0.307 
			&      0.641 &      0.639 
			&      0.347 &      0.417 
			&      0.301 &      0.319 
			&      0.330 &      0.401 
			&      0.291 &      0.381 
			&      0.885 &      0.711 
			\\
			\midrule

			\multirow{5}{*}{\rotatebox{90}{PEMS08}}
			&         12 &\txbl{0.077}&\txbl{0.180}
			&\txrd{0.076}&\txrd{0.178}
			&      0.079 &      0.182 
			&      0.133 &      0.247 
			&      0.168 &      0.232 
			&      0.165 &      0.214 
			&      0.227 &      0.343 
			&      0.112 &      0.212 
			&      0.154 &      0.276 
			&      0.173 &      0.273 
			&      0.436 &      0.485 
			\\

			&         24 &      0.117 & 0.222 
			&\txrd{0.104}&\txrd{0.209}
			&\txbl{0.115}&\txbl{0.219}
			&      0.249 &      0.343 
			&      0.224 &      0.281 
			&      0.215 &      0.260 
			&      0.318 &      0.409 
			&      0.141 &      0.238 
			&      0.248 &      0.353 
			&      0.210 &      0.301 
			&      0.467 &      0.502 
			\\

			&         48 &      0.212 & 0.302 
			&\txrd{0.167}&\txrd{0.228}
			&\txbl{0.186}&\txbl{0.235}
			&      0.569 &      0.544 
			&      0.321 &      0.354 
			&      0.315 &      0.355 
			&      0.497 &      0.510 
			&      0.198 &      0.283 
			&      0.440 &      0.470 
			&      0.320 &      0.394 
			&      0.966 &      0.733 
			\\

			&         96 &      0.306 &      0.354 
			&\txbl{0.245}&\txbl{0.280}
			&\txrd{0.221}&\txrd{0.267}
			&      1.166 &      0.814 
			&      0.408 &      0.417 
			&      0.377 &      0.397 
			&      0.721 &      0.592 
			&      0.320 &      0.351 
			&      0.674 &      0.565 
			&      0.442 &      0.465 
			&      1.385 &      0.915 
			\\

			\cmidrule(lr){2-24}
			&        Avg &     {0.178}& {0.265} 
			&\txrd{0.148}&\txrd{0.224}
			&\txbl{0.150}&\txbl{0.226}
			&      0.529 &      0.487 
			&      0.280 &      0.321 
			&      0.268 &      0.307 
			&      0.441 &      0.464 
			&      0.193 &      0.271 
			&      0.379 &      0.416 
			&      0.286 &      0.358 
			&      0.814 &      0.659 
			\\
			\bottomrule
		\end{tabular}
	\end{small}
}
\vspace{-0.1in}
\end{table*}


\subsection{Ablation Study}
To assess the contribution of each component in \textbf{MODE}, we conduct ablation studies summarized in Table~\ref{tab:ablation}. We analyze: (1) the use of Neural ODEs, (2) dynamic vs. static low-rank parameterization, (3) the effect of rank size $r$, and (4) the interaction between the ODE formulation and the Mamba architecture. Experiments are run on ETTm1, Weather, and ECL with forecast horizons $T \in \{96, 192, 336, 720\}$, using evaluating metrics MSE and MAE.

\textbf{Effect of Low-Rank Neural ODEs.} \xs{Compared with vanilla Mamba and bidirectional Mamba blocks in parallel, Neural ODE variants are on par with or outperform models without ODE implementations across datasets and horizons. For example, at horizon 96 on ETTm1, \textbf{\model} with static low-rank ODE achieves an MSE of \textbf{0.324}, outperforming vanilla Mamba (0.326) and bidirectional Mamba (0.331) and is close to attention mechanism. It also achieves best or second best performances in horizon 336 and 720 experiments. This shows that continuous-time dynamics help capture irregular temporal patterns that discrete models miss.}

\textbf{Dynamic versus Static Low-Rank Parameterization.} The dynamic variant further improves over the static one. On Weather at horizon 96, Dynamic Low-Rank ODE achieves the best MSE/MAE (0.167/0.210), surpassing the static ODE (0.170/0.213). \xs{This suggests that, at the expense of some time and memory efficiency, adaptive ranks better capture time-varying dependencies, improving stability and accuracy.}

\textbf{Impact of Rank Size.} We vary the rank ratio $\frac{r}{d_{\text{state}}}$ from 1/4 to 1/2 and full. Moderate ranks (1/2 $d_{\text{state}}$) offer the best efficiency–performance trade-off. Full rank yields only marginal accuracy gains while increasing cost, supporting the choice of a low-rank design.

\xs{\textbf{Influence of Architectural Components.} We also test variants that replace Mamba enhanced by low rank ODE with vanilla Mamba, S-Mamba block, attention + FFN, or pure linear layers. Attention solution shows advantages in short-term forecasting, for its ability to capture high correlation between future and recent values, and respond to local variations by assigning higher weights to nearby timesteps. But \model~shows competitive performance with a considerable number of best or second best error results in all prediction length, indicating that both the ODE formulation and its integration into the Mamba pipeline are key to \textbf{\model}'s improvements.}

Overall, the ablations show that: (1) Neural ODEs improve temporal modeling, (2) dynamic low-rank parameterization enhances adaptability and generalization, and (3) the synergy between Neural ODEs and Mamba’s selective scanning drives both accuracy and efficiency, validating the design of each module.


\begin{table*}[t]
	\centering
	\footnotesize
	\caption{Ablation Study Results. The subtable on the left shows prediction results with varied ODE implementations or other types of block. The subtable on the right shows results given different ratios between $r$ and $d_{state}$.}
	\label{tab:ablation}

	\begin{minipage}{0.45\textwidth}
	\centering
	\setlength{\tabcolsep}{3pt}
	\begin{tabular}{l|c|c|c|c|c|c|c|c}
    \toprule
	\rule{0pt}{4pt}
	\multirow{2}{*}{Model} &
	\multirow{2}{*}{R Rank} &
	\multirow{2}{*}{Len} &
	\multicolumn{2}{c|}{ETTm1} &
	\multicolumn{2}{c|}{Weather} &
	\multicolumn{2}{c}{ECL} \\
	\cline{4-5}\cline{6-7}\cline{8-9}
	\rule{0pt}{8pt}& & & MSE & MAE & MSE & MAE & MSE & MAE \\
	\midrule

	\multirow{4}{*}{\parbox{1.4cm}{Static ODE}} & \multirow{4}{*}{Custom}
		& 96    &      0.324 &      0.363 &      0.170 &0.213&      0.150 &      0.244 \\
		& & 192 &      0.364 &      0.384 &      0.219 &0.255&\txrd{0.164}&      0.256 \\
		& & 336 &\txrd{0.398}&\txrd{0.407}&\txrd{0.274}&0.296&\txrd{0.180}&\txrd{0.272}\\
		& & 720 &      0.468 &      0.447 &      0.354 &0.348&     {0.214}&\txrd{0.302}\\
	\cmidrule{1-9}
	\multirow{4}{*}{\parbox{1.4cm}{Dynamic ODE}} & \multirow{4}{*}{Custom}
		&    96 &  0.332  &  0.366  &       {0.167} &      {0.210} &  0.163 &  0.254 \\
		& & 192 &  0.367  &  0.385  &       {0.216} & \txrd{0.253} &  0.172 &  0.261 \\
		& & 336 &  0.401  &  0.408  &  \txrd{0.274} & \txrd{0.295} &  0.190 &  0.279 \\
		& & 720 &  0.477  &  0.453  &  \txrd{0.353} & \txrd{0.346} &  0.231 &  0.313 \\
	\cmidrule{1-9}
	\multirow{4}{*}{\parbox{1.4cm}{Vanilla Mamba}} & \multirow{4}{*}{Custom}
		& 96    &  0.326  &  0.364  &  0.169  &  0.212  &      {0.150} &      0.245 \\
		& & 192 &  0.367  &  0.386  &  0.219  &  0.256  & \txrd{0.164} &     {0.256}\\
		& & 336 &  0.401  &  0.409  &  0.275  &  0.297  & \txrd{0.180} &      0.273 \\
		& & 720 &  0.482  &  0.455  &  0.354  &  0.349  &      {0.214} &\txrd{0.302}\\
	\cmidrule{1-9}
	\multirow{4}{*}{\parbox{1.6cm}{Bidirectional Mamba}} & \multirow{4}{*}{\scriptsize Custom}
		&   96&0.331&0.367&\txrd{0.166}&\txrd{0.209}&\txrd{0.143}&\txrd{0.239}\\
		& &192&0.377&0.390&\txrd{0.215}&      0.254 &\txrd{0.164}&      0.259 \\
		& &336&0.412&0.416&      0.277 &      0.296 &     {0.190}&      0.283 \\
		& &720&0.485&0.457&      0.355 &      0.350 &\txrd{0.213}&      0.304 \\
	\cmidrule{1-9}
	\multirow{4}{*}{\parbox{1.4cm}{Attention + FFN}} & \multirow{4}{*}{Custom}
		&   96&\txrd{0.320}&\txrd{0.357}&0.175&0.215&      0.152 &     {0.243}\\
		& &192&\txrd{0.363}&\txrd{0.380}&0.223&0.258&\txrd{0.164}&\txrd{0.254}\\
		& &336&      0.405 &      0.409 &0.281&0.300&\txrd{0.180}&\txrd{0.272}\\
		& &720&\txrd{0.467}&\txrd{0.445}&0.360&0.350&      0.221 &      0.306 \\
	\cmidrule{1-9}
	\multirow{4}{*}{Linear} & \multirow{4}{*}{Custom}
		& 96    &  0.342  &  0.365  &  0.190  &  0.230  &  0.196  &  0.273  \\
		& & 192 &  0.386  &  0.388  &  0.235  &  0.266  &  0.196  &  0.275  \\
		& & 336 &  0.427  &  0.416  &  0.288  &  0.304  &  0.211  &  0.291  \\
		& & 720 &  0.486  &  0.447  &  0.362  &  0.353  &  0.253  &  0.324  \\
	\bottomrule
	\end{tabular}
	\end{minipage}
	\hspace{35pt}
	\begin{minipage}{0.45\textwidth}
	\centering
	\setlength{\tabcolsep}{3pt}
	\begin{tabular}{c|c|c|c|c|c|c|c|c}
	\toprule
	\rule{0pt}{4pt}
	\multirow{2}{*}{ODE} &
	\multirow{2}{*}{R Rank} &
	\multirow{2}{*}{Len} &
	\multicolumn{2}{c|}{ETTm1} &
	\multicolumn{2}{c|}{Weather} &
	\multicolumn{2}{c}{ECL} \\
	\cline{4-5}\cline{6-7}\cline{8-9}
	\rule{0pt}{8pt}& & & MSE & MAE & MSE & MAE & MSE & MAE \\
	\midrule

	& \multirow{4}{*}{\scriptsize 1/4 $d_{\text{state}}$}
		& 96    &      0.330 &      0.366 &0.170&0.213&\txrd{0.150}&      0.245 \\
		& & 192 &      0.370 &      0.388 &0.218&0.255&\txrd{0.164}&\txrd{0.256}\\
		& & 336 &\txrd{0.398}&\txrd{0.407}&0.275&0.297&      0.181 &      0.273 \\
		& & 720 &      0.477 &      0.452 &0.353&0.348&\txrd{0.216}&\txrd{0.304}\\
	\cmidrule{2-9}
	\multirow{4}{*}{\rotatebox{90}{Static}} & \multirow{4}{*}{\scriptsize 1/2 $d_{\text{state}}$}
		& 96    &\txrd{0.324}&\txrd{0.363}&0.171&0.214&\txrd{0.150}&\txrd{0.244}\\
		& & 192 &      0.364 &      0.384 &0.219&0.255&\txrd{0.164}&      0.257 \\
		& & 336 &      0.399 &      0.408 &0.275&0.296&      0.181 &      0.273 \\
		& & 720 &      0.474 &      0.450 &0.353&0.348&      0.217 &\txrd{0.304}\\
	\cmidrule{2-9}
	& \multirow{4}{*}{\scriptsize $d_{\text{state}}$}
		& 96    &      0.329 &      0.365 &0.170&0.213&\txrd{0.150}&\txrd{0.244}\\
		& & 192 &\txrd{0.363}&\txrd{0.382}&0.218&0.255&\txrd{0.164}&\txrd{0.256}\\
		& & 336 &      0.401 &      0.409 &0.275&0.296&\txrd{0.180}&\txrd{0.272} \\
		& & 720 &      0.480 &      0.454 &0.353&0.347&      0.217 &\txrd{0.304} \\
	\cmidrule{1-9}
	& \multirow{4}{*}{\scriptsize 1/4 $d_{\text{state}}$}
		& 96    &  0.328  &\txrd{0.363}&      0.168 &     {0.211}&  0.163  &  0.253 \\
		& & 192 &  0.368  &      0.385 &      0.217 &\txrd{0.253}&  0.173  &  0.261 \\
		& & 336 &  0.402  &      0.408 &\txrd{0.274}&\txrd{0.295}&  0.190  &  0.279 \\
		& & 720 &  0.473  &      0.451 &\txrd{0.352}&\txrd{0.346}&  0.229  &  0.311 \\
	\cmidrule{2-9}
	\multirow{4}{*}{\rotatebox{90}{Dynamic}} & \multirow{4}{*}{\scriptsize 1/2 $d_{\text{state}}$}
		& 96    &       0.332  &0.366&\txrd{0.167}&\txrd{0.210}&  0.163  &  0.252 \\
		& & 192 &       0.367  &0.385&\txrd{0.216}&\txrd{0.253}&  0.173  &  0.263 \\
		& & 336 &       0.401  &0.408&\txrd{0.274}&\txrd{0.295}&  0.190  &  0.279 \\
		& & 720 & \txrd{0.465} &0.446&     {0.353}&\txrd{0.346}&  0.230  &  0.314 \\
	\cmidrule{2-9}
	& \multirow{4}{*}{\scriptsize $d_{\text{state}}$}
		& 96    &0.328&      0.364 &0.168&      0.212 &0.163&0.253 \\
		& & 192 &0.369&      0.387 &0.217&      0.254 &0.173&0.262 \\
		& & 336 &0.402&      0.408 &0.275&\txrd{0.295}&0.190&0.280 \\
		& & 720 &0.467&\txrd{0.444}&0.353&\txrd{0.346}&0.229&0.311 \\
	\bottomrule
	\end{tabular}
	\end{minipage}
\end{table*}


\begin{figure*}[!t]
    \centering
    \includegraphics[width=\textwidth]{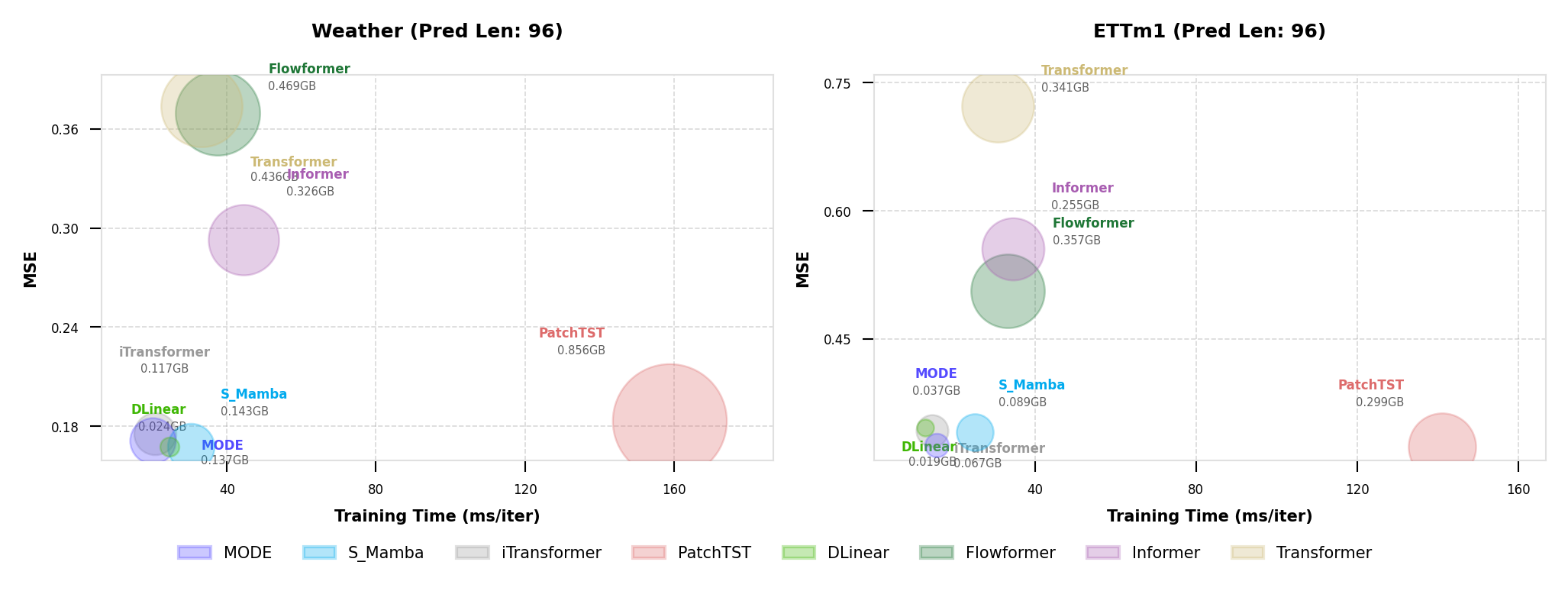}
    \caption{Efficiency comparison of our method \model\ and baselines on four datasets in terms of training time.}
    \label{fig:efficiency}
\end{figure*}

\subsection{Efficiency Comparison}

Fig.~\ref{fig:efficiency} illustrates the efficiency comparison between our proposed \textbf{\model} and several strong baselines, including S-Mamba, PatchTST, iTransformer, Flowformer, Informer, \xs{Transformer, and DLinear, on the Weather and ETTm1 datasets. The x-axis represents the average training time per iteration, the y-axis shows MSE results, and the bubble size indicates allocated GPU memory under same batching settings (128 for PatchTST, 16 for other models). \textbf{\model} with static low-rank ODE achieves an excellent balance between accuracy and efficiency, yielding lower MSE with competitive training time and small memory footprint. Specifically, \textbf{\model} achieves comparable or better accuracy than PatchTST and iTransformer while requiring significantly less training time and reduced memory usage (only 0.037~GB on ETTm1 and 0.137~GB on Weather). In contrast, Transformer-based and Flowformer architectures incur higher computational costs and memory overhead. These results demonstrate that the low-rank Neural ODE formulation and selective scanning mechanism enable \textbf{\model} to achieve both high predictive accuracy and superior computational efficiency.}

\subsection{Robustness Study}
\begin{figure*}[!t]
    \centering
    \includegraphics[width=\textwidth]{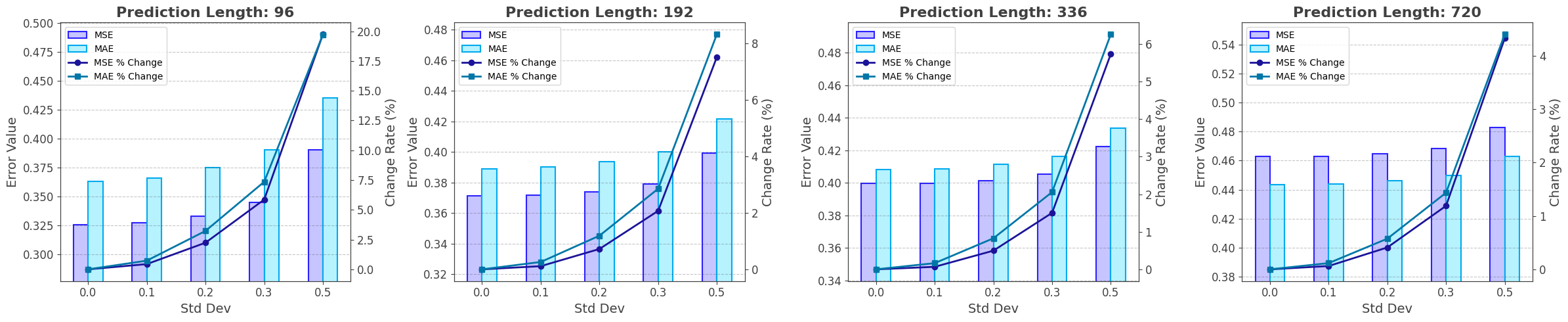}
    \caption{Robustness experiments of \model\ on ETTm1 with varied standard deviations of gaussian noise.}
    \label{fig:robustness}
\end{figure*}

To evaluate the robustness of \textbf{\model} against input perturbations, we conduct experiments on the ETTm1 dataset by injecting Gaussian noise with varying standard deviations into the input sequences. Fig.~\ref{fig:robustness} presents the results under four forecasting horizons $T \in \{96, 192, 336, 720\}$. The left axis reports the absolute error values (MSE and MAE), while the right axis shows the relative performance degradation as noise intensity increases. \xs{As illustrated, \textbf{\model} maintains stable performance under low to moderate noise levels (standard deviation $\leq 0.3$), with only marginal increases in MSE and MAE. Even under heavy noise injection ($\text{std}=0.5$), the error growth is accelerated but remains relatively moderate. Except for horizon 96 experiments, MSE and MAE growths are consistently kept within 10\%}. This robustness arises from the continuous-time dynamics of the Neural ODE formulation, which smooths noisy variations, and the selective scanning mechanism, which emphasizes informative temporal segments. These results confirm that \textbf{\model} is resilient and reliable in noisy and uncertain environments.

\begin{figure*}[!t]
    \centering
    \includegraphics[width=\textwidth]{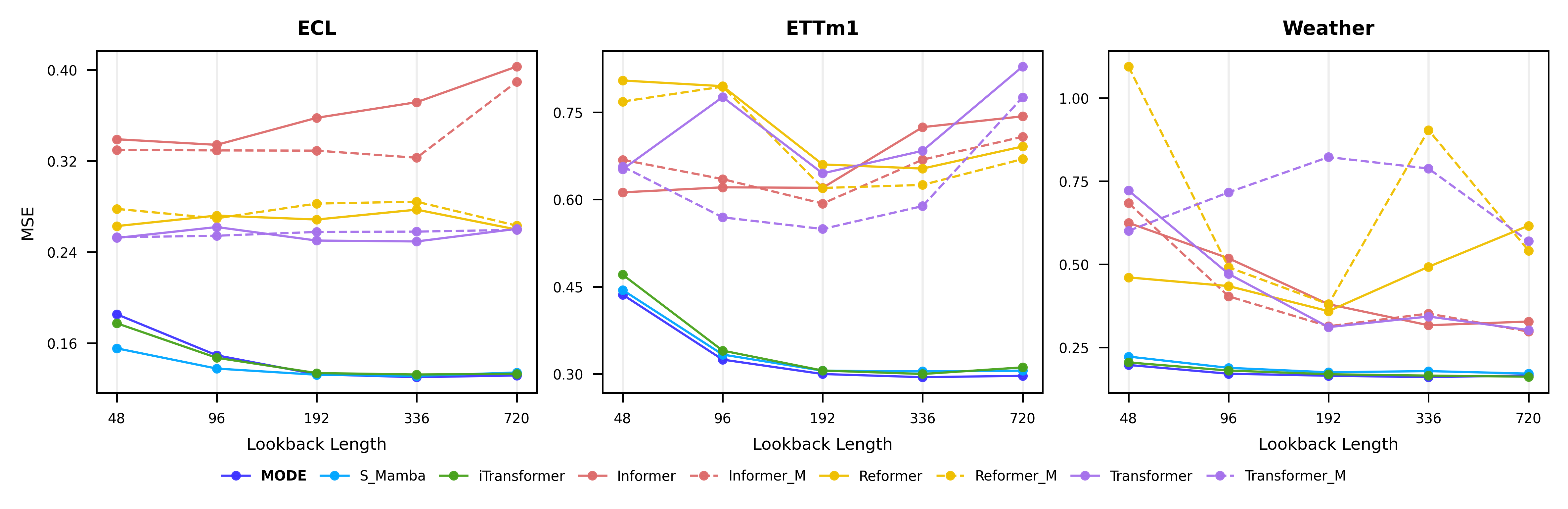}
    \caption{Long-term prediction comparison experiments of \model\ on datasets with increasing lookback length.}
    \label{fig:lookback}
\end{figure*}

\subsection{Long-term Prediction Comparison}
To assess the capability of \textbf{\model} in modeling long-range temporal dependencies, we conduct experiments with varying lookback window lengths on the ECL, ETTm1, and Weather datasets. Fig.~\ref{fig:lookback} presents the MSE results as the lookback length increases from 48 to 720. As shown, \textbf{\model} consistently achieves the lowest or near-lowest error across all datasets and lookback settings, demonstrating strong adaptability to long-term dependency modeling. Unlike Transformer-based and Reformer-based approaches, whose performance tends to degrade or fluctuate at large lookback horizons due to attention saturation or inefficient memory usage, \textbf{\model} maintains stable and robust accuracy. This improvement stems from the integration of low-rank Neural ODEs, which capture smooth and continuous temporal dynamics, and the selective scanning mechanism, which focuses computation on informative subsequences. Overall, these results validate that \textbf{\model} effectively leverages longer historical contexts for stable and accurate long-term forecasting across diverse time series domains.

\subsection{Case Study}

\begin{figure*}[!t]
    \centering
    \includegraphics[width=\textwidth]{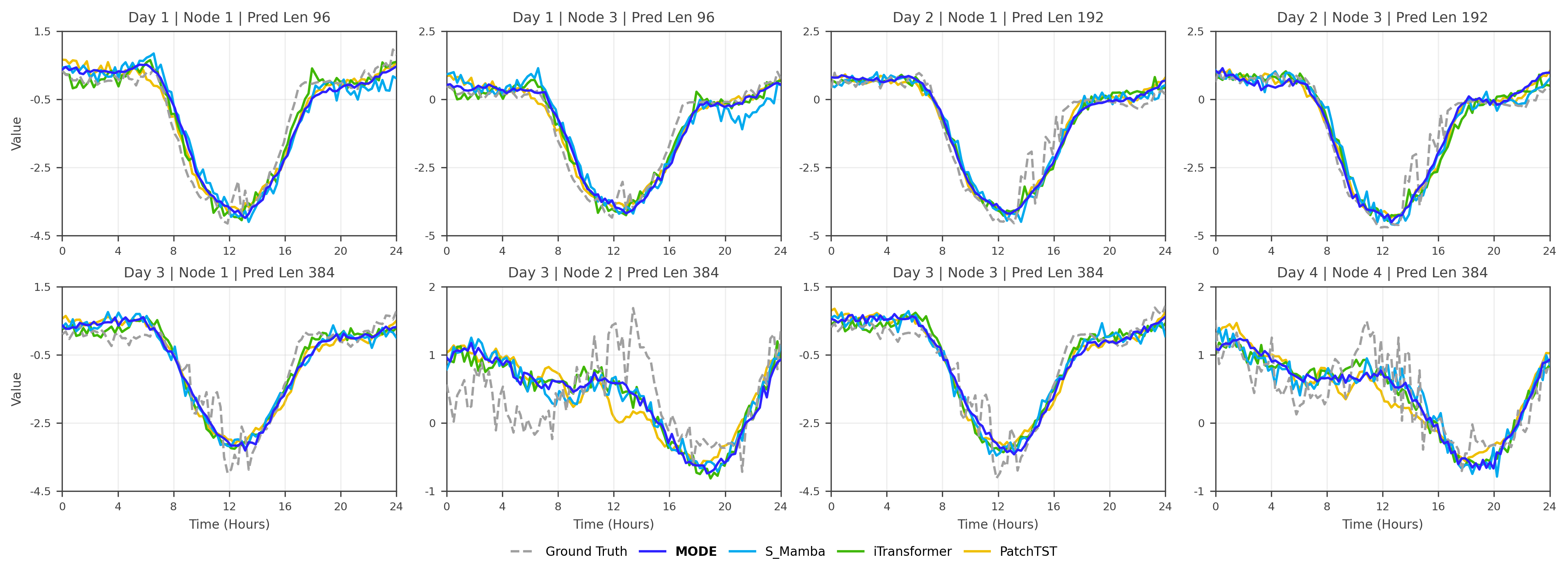}
    \caption{Case study of \textbf{\model} on ETTm1 across different horizons and scenarios.}
    \label{fig:case_study}
\end{figure*}

To provide intuitive insights into the effectiveness of our proposed \model\ method, we conduct a detailed case study analysis on the ETTm1 dataset across multiple prediction horizons and scenarios. Fig.~\ref{fig:case_study} presents representative forecasting results comparing \model\ against baseline methods including S-Mamba, iTransformer, and PatchTST across different days and prediction lengths.

The case study reveals several key observations that highlight \model\ 's superior performance:

\textbf{Accurate Trend Capture}: Across all scenarios, \model\ (purple solid line) consistently follows the ground truth patterns more closely than baseline methods. This is particularly evident in the U-shaped temperature curves observed throughout different days, where MODE accurately captures both the magnitude and timing of temperature variations.

\textbf{Robust Long-term Forecasting}: As the prediction horizon extends from 96 to 384 time steps, \model\ maintains stable performance while baseline methods show increasing deviation. For instance, in Day 3 with 384-step prediction, competing methods exhibit significant oscillations and drift, whereas \model\ preserves the underlying trend structure.

\textbf{Superior Handling of Complex Patterns}: MODE demonstrates exceptional capability in modeling intricate temporal dependencies, especially during transition periods where temperature changes rapidly. The method successfully reconstructs the smooth temperature curves while avoiding the erratic fluctuations observed in baseline approaches.

\textbf{Consistency Across Different Scenarios}: The case study encompasses various challenging conditions including different seasonal patterns (Day 1 vs Day 4) and varying prediction complexities. \model\ consistently outperforms alternatives, demonstrating its robustness and generalization capability across diverse forecasting scenarios.

These qualitative results complement our quantitative analysis and provide compelling evidence of \model\ 's effectiveness in capturing complex temporal dynamics for accurate time series forecasting.

\section{Related Work}

Modern time series forecasting has progressed rapidly with the development of deep sequence models, particularly Transformer-based architectures~\cite{vaswani2017attention}. These models capture long-range temporal dependencies through the self-attention mechanism, enabling accurate multistep forecasting. Extensions of the Transformer framework~\cite{lim2021time,torres2021deep,zheng2021graph} incorporate causal masking to preserve the temporal order, while hierarchical designs such as Pyraformer~\cite{liu2021pyraformer} and cross-variate modules like Crossformer~\cite{zhang2022crossformer} further enhance contextual representation. However, their quadratic time complexity $\mathcal{O}(N^2)$ and memory cost limit scalability to long sequences. To address this, numerous studies have explored linearized or window-based attention mechanisms, yet these often trade off precision for efficiency. Large-scale pretraining frameworks such as Moirai~\cite{woo2024unified} offer another direction but differ substantially in setup and scale from our controlled evaluation.

Recently, Mamba-based architectures~\cite{gu2023mamba,zhang2025hmamba,zhang2025fldmamba,zhang2025m2rec,zhang2025autohformer} have emerged as promising state-space models, demonstrating strong performance in temporal reasoning and long-context sequence modeling. While these methods improve computational efficiency, they remain limited in handling irregularly sampled sequences and continuous-time dynamics.

In contrast, our proposed \textbf{\model} introduces the first unified framework that combines Low-Rank Neural Ordinary Differential Equations (Neural ODEs) with an Enhanced Mamba architecture. This integration enables continuous-time modeling, efficient long-range dependency capture, and adaptive computational allocation through selective scanning. By bridging the gap between ODE-based temporal modeling and scalable state-space architectures, \textbf{\model} achieves state-of-the-art accuracy while maintaining superior efficiency and robustness compared to prior Transformer and Mamba variants.

\section{Conclusion}

In this work, we present \textbf{\model}, a unified and efficient framework for time series forecasting that integrates Low-Rank Neural Ordinary Differential Equations (Neural ODEs) with the Mamba architecture. By combining the continuous-time modeling capabilities of Neural ODEs with Mamba’s selective scanning mechanism, \textbf{\model} effectively captures long-range temporal dependencies, gracefully handles irregular sampling patterns, and substantially decreases computational overhead. The proposed low-rank parameterization further improves scalability by reducing the state-transition complexity from $\mathcal{O}(d^2)$ to $\mathcal{O}(d \cdot r)$, while preserving rich representational power. Extensive experiments on diverse benchmark datasets show that \textbf{\model} consistently delivers state-of-the-art accuracy, efficiency, and robustness. Comprehensive ablation and robustness analyses validate each architectural component and confirm resilience to noise and long forecasting horizons. Future directions include integrating probabilistic inference and physics-informed priors for enhanced uncertainty quantification and broader real-world applicability.

\bibliographystyle{plain}

\bibliography{ref}

\end{document}